\documentclass[letterpaper, 10pt, conference]{ieeeconf}
\IEEEoverridecommandlockouts

\usepackage{amsthm}
\usepackage{algorithmic}
\usepackage{xcolor}
\usepackage[utf8]{inputenc}
\usepackage{amsmath}
\usepackage{empheq}
\usepackage{amsfonts}
\usepackage{bbold}
\usepackage{booktabs}
\usepackage{tikz}
\usepackage{siunitx}
\usepackage{float}
\usepackage[noadjust]{cite}
\usepackage[colorlinks,allcolors=black]{hyperref}
\usepackage{graphicx}
\graphicspath{ {./images/} }
\usepackage[english]{babel}
\usepackage{pgfplots}
\usepackage{balance}
\usepackage{amssymb}
\usepackage{mathtools}

\usetikzlibrary{shapes.geometric}

\usepgfplotslibrary{fillbetween}
\pgfplotsset{compat=1.12}
\pgfdeclarelayer{ft}
\pgfdeclarelayer{bg}
\pgfsetlayers{bg,main,ft}

\DeclareMathSymbol{\shortminus}{\mathbin}{AMSa}{"39}

\theoremstyle{plain}
\newtheorem{definition}{Definition}

\def\BibTeX{{\rm B\kern-.05em{\sc i\kern-.025em b}\kern-.08em
    T\kern-.1667em\lower.7ex\hbox{E}\kern-.125emX}}
\begin{document}

\author{Nishanth Rao$^{1}$, Suresh Sundaram$^{1,2}$, and Pushpak Jagtap$^{2}$
\thanks{$^{1}$Pushpak Jagtap is with the Robert Bosch Center for Cyber-Physical Systems, Nishanth Rao is with the Department of Aerospace Engineering, and Suresh Sundaram is with the Robert Bosch Center for Cyber-Physical Systems and the Department of Aerospace Engineering at Indian Institute of Science, Bangalore, India
        {\tt\small nishanthrao,vssuresh,pushpak@iisc.ac.in}}%
}

\title{Temporal Waypoint Navigation of Multi-UAV Payload System using Barrier Functions}

\maketitle

\begin{abstract}
Aerial package transportation often requires complex spatial and temporal specifications to be satisfied in order to ensure safe and timely delivery from one point to another. It is usually efficient to transport versatile payloads using multiple UAVs that can work collaboratively to achieve the desired task. The complex temporal specifications can be handled coherently by applying Signal Temporal Logic (STL) to dynamical systems. This paper addresses the problem of waypoint navigation of a multi-UAV payload system under temporal specifications using higher-order time-varying control barrier functions (HOCBFs). The complex nonlinear system of relative degree two is transformed into a simple linear system using input-output feedback linearization. An optimization-based control law is then derived to achieve the temporal waypoint navigation of the payload. The controller's efficacy and real-time implementability are demonstrated by simulating a package delivery scenario inside a high-fidelity \texttt{Gazebo} simulation environment.
\end{abstract}

\section{Introduction}
Advancements in multi-UAV system research and technologies have led to the widespread use of multi-UAV systems in the aerial transportation of payloads from one point to another. It is evident that using multiple UAVs to deliver versatile payloads is more efficient than utilizing a single UAV of varying lift capacities. In addition, payloads need to be delivered to specific locations under some time constraints. For example, in the air delivery of packages, it is crucial to not only deliver the package at the correct location, but the package must also reach within the specified time. Thus it is essential to design control algorithms for multi-UAV payload systems that follow certain \textit{temporal} specifications while enabling safe payload transportation.

Most of the literature focuses on developing trajectory-tracking controllers for multi-UAV payloads without considering temporal specifications. In \cite{lee2017geometric}, the authors propose a trajectory tracking controller using geometric control for a UAV fleet carrying the payload using cable suspension. In \cite{wehbeh2020distributed}, a distributed model predictive control algorithm is developed for the purpose of trajectory tracking of payload using rigid link payload suspension. 
The authors in \cite{9945662} develop an adaptive controller for a cable-suspended payload with user-specified safety constraints. For a rigid link suspended payload, a feedback linearized nonlinear controller is presented in \cite{rao2022input}.

Tasks that require complex spatial and temporal requirements can be handled effectively by applying Signal Temporal Logic (STL) \cite{maler2004monitoring} to dynamical systems. The authors in \cite{lindemann2018control} use control barrier functions \cite{ames2016control, ames2019control} to design control inputs enforcing a class of STL specifications. In \cite{pant2021co}, a dynamically feasible STL-based planning and control algorithm is designed for a single UAV system that satisfies complex spatial and temporal requirements. An extension to multi-UAV fleets (without payload) is proposed in \cite{pant2018fly} that generates continuous trajectories for multiple UAVs to meet the temporal requirements.   

While there have been works on designing controllers for multi-agent systems to satisfy STL specifications \cite{9161270}, there has been no work on developing controllers to satisfy complex STL specifications, particularly for a multi-UAV system carrying a payload. One of the reasons is that the dynamics of such a system is highly coupled and nonlinear in nature. This can lead to a complex state-space representation with large state and control vector dimensions. The \textit{curse of dimensionality} can thus make it difficult to control such a system under STL specifications.

This paper focuses on developing a controller for the multi-UAV payload system to enforce the complex STL specifications using \textit{Higher-Order Time-Varying Control Barrier Functions} (HOCBFs). The model complexity of the system is reduced by the application of \textit{input-output feedback linearization}, which results in a linear system with a \textit{relative degree} equal to 2. The STL specifications are captured using time-varying HOCBFs, and an optimization-based control law is proposed. To validate the efficacy and real-time implementability of the proposed controller, a simulation environment is developed using the \texttt{Gazebo} simulator for the purpose of simulating a package delivery scenario under time specifications.

Section \ref{sec:prelim} discusses the necessary preliminaries and notations used in this paper. Section \ref{sec:TempNav} proposes a control law for \textit{temporal waypoint navigation} of the multi-UAV payload system. In Section \ref{sec:results}, the proposed control law is validated in a high-fidelity \texttt{Gazebo} simulator that simulates a package-delivery scenario. Section \ref{sec:future} concludes the proposed work.

\section{Preliminaries}
\label{sec:prelim}
In this paper, scalars and vectors are denoted by non-bold letters $a$, whereas matrices are denoted by bold letters $\pmb{A}$. We use $\mathbb{R}$, $\mathbb{R}^+$, and $\mathbb{R}^+_0$ to denote real, positive real, and nonnegative real numbers, respectively. $\mathbb{R}^n$ denotes an $n$-dimensional vector with real elements and $\mathbb{R}^{m\times n}$ denotes an $m\times n$ matrix with real elements. The vector $\mathbb{k} = \left[ 0 \ 0 \ 1 \right]^T \in \mathbb{R}^3$ represents the \textit{unit} $z$-axis. The matrix $\pmb{I}_n\in\mathbb{R}^{n\times n}$ denotes the $n\times n$ \textit{identity} matrix. The \textit{Moore-Penrose pseudoinverse} of a general matrix $A\in\mathbb{R}^{m\times n}$ is denoted by $A^\dagger\in\mathbb{R}^{n\times m}$. A real-valued $n$-dimensional signal is denoted by $x:\mathbb{R}^+_0 \rightarrow \mathbb{R}^n$. The notation $x$ is abused by also referring to the trajectory of a dynamical system with $n$ states. Thus, the meaning of $x$ is assumed to be clear from the context. A function $\alpha:\mathbb{R}_0^+ \rightarrow \mathbb{R}_0^+$ is a class $\mathcal{K}$ function if it is continuous and strictly increasing, with $\alpha(0)=0$. The \textit{lie derivative} of a vector field $b$ along the vector field $a$ is denoted by $\mathcal{L}_ab = a \cdot \frac{\partial b}{\partial x}$. The function composition operator $\circ$ is defined as $f\circ g = f(g(.))$.  

\subsection{Payload-UAV System Description}
\label{subsec:desc}
There are, in general, $N$ UAVs connected to the rigid massless rods, each of length $l_i$ via \textit{spherical joints}. These rods are rigidly connected to the payload and thus always remain vertical. The location of the center of mass of the payload and the $i^{th}$ UAV is denoted by $r_0 \in \mathbb{R}^3$ and $r_i \in \mathbb{R}^3$ respectively. The vector $\rho_i \in \mathbb{R}^3$ denotes the position of the attachment between the rod corresponding to the $i^{th}$ UAV and the payload with respect to the center of mass of the payload. The mass of the payload and the $i^{th}$ UAV is denoted by $m_0, m_i \in \mathbb{R}^+$ respectively. The inertia matrix of the payload and the $i^{th}$ UAV is indicated by $\pmb{J}_0, \pmb{J}_i \in \mathbb{R}^{3\times 3}$ respectively.  The rotation matrix $\pmb{R}_0 \in SO(3)$ and $\pmb{R}_i \in SO(3)$ denotes the orientation of the payload and the $i^{th}$ UAV respectively. This rotation matrix converts a vector in the body-fixed frame to the inertial frame. The payload-UAV system is modelled in the standard North, East and Down frame (NED), with the positive unit $z$-axis pointing downwards.
The total thrust force produced by all the motors of the $i^{th}$ UAV is denoted by $F_{t_i} \in \mathbb{R}^+$ and is related to the force vector $F_i \in \mathbb{R}^3$ as $F_i = F_{t_i}\pmb{R}_i\mathbb{k}$. The motors produce a torque vector in the body frame of the $i^{th}$ UAV denoted by $\tau_i = \left[\tau_{x_i} \ \tau_{y_i} \ \tau_{z_i}\right]^T \in \mathbb{R}^3$. Thus, the overall control input for the multi-UAV system is $\{F_{t_i}, \tau_i\}$ with $i = \left\{1, \hdots, N\right\}$.

\subsection{Payload-UAV Dynamics and State-space Representation}
Using the description of the payload-UAV system discussed in Section \ref{subsec:desc}, the equations of motion for the multi-UAV payload system (shown in Fig. \ref{fig:schematic_diagram_payloadUAV}) can be derived. As the rigid links are always vertical, the position of the $i^{th}$ UAV can be inferred as:
\begin{align}
    r_i = r_0 + \pmb{R}_0\left(\rho_i - l_i\mathbb{k} \right).
\end{align}
The kinematic equations are given by:
\begin{align}
    \dot{r}_0 & = \pmb{R}_0v_0, \label{eq:kin_beg}\\
    \dot{\pmb{R}}_0 & = \pmb{R}_0\omega_0^\times ,\\
    \dot{r}_i & = \dot{r}_0 + \pmb{R}_0\omega_0^\times \left(\rho_i - l_i\mathbb{k} \right), \\
    \dot{\pmb{R}}_i & = \pmb{R}_i\omega_i^\times, \label{eq:kin_end}
\end{align}
where the skew-symmetric operator $(\cdot)^\times: \mathbb{R}^3 \rightarrow SO(3)$ denotes the \textit{hat map} which maps a vector to a skew-symmetric matrix, $v_0 \in \mathbb{R}^3$ denotes the linear velocity of the payload in the payload-fixed frame, $\omega_0\in\mathbb{R}^3$ and $\omega_i\in\mathbb{R}^3$ denotes the angular velocities of the payload and the $i^{th}$ UAV, respectively.

\begin{figure}
    \begin{tikzpicture}
    
        \node[name path=A, trapezium, draw, minimum width=3.5cm, trapezium left angle=120, trapezium right angle=60, line width=0mm, fill=brown!15] at (2.5,-3) {};
        \node(Y)[name path=B, trapezium, draw, minimum width=4cm, trapezium left angle=120, trapezium right angle=60, line width=0mm] at (2.5,-3) {};
        
        \draw (3.3, -2.7) -- (3.3, -2.2);
        \draw (1.05, -2.7) -- (3.3, -2.7) -- (3.95, -3.8);
        
        \draw[name path=C] (0.5, -2.08) -- (1.57, -3.93) -- (4.5, -3.93);
        \draw[name path=D] (0.5, -2.8) -- (1.57, -4.5) -- (4.5, -4.5);
        
        \draw (0.5, -2.08) -- (0.5, -2.8);
        \draw (1.57, -4.5) -- (1.57, -3.93);
        \draw (4.5, -4.5) -- (4.5, -3.93);
        
        \tikzfillbetween[of=A and B, on layer=ft]{brown, opacity=0.5};
        \tikzfillbetween[of=C and D, on layer=ft]{brown, opacity=0.5};
        
        \draw[name path=E] (0.55, -2.9) -- (0.55, 1.0); 
        \draw[name path=F] (0.6, -3.0) -- (0.6, 0.95);
        \tikzfillbetween[of=E and F, on layer=ft]{black, opacity=0.7};
        
        \draw[name path=G] (1.5, -4.4) -- (1.5, -1.0);
        \draw[name path=H] (1.45, -4.3) -- (1.45, -0.95);
        \tikzfillbetween[of=G and H, on layer=ft]{black, opacity=0.7};
        
        \draw[name path=I] (4.5, -3.95) -- (4.5, -1.0);
        \draw[name path=J] (4.45, -3.85) -- (4.45, -0.95);
        \tikzfillbetween[of=I and J, on layer=ft]{black, opacity=0.7};
    
        \draw[name path=K] (3.55, -2.3) -- (3.55,0.95);
        \draw[name path=L] (3.5, -2.2) -- (3.5, 1.0);
        \tikzfillbetween[of=K and L, on layer=ft]{black, opacity=0.7};
        
        \node[cylinder, draw=black!40, shape border rotate=90, minimum width=1.7cm, minimum height=0.8cm, cylinder uses custom fill, cylinder body fill = black!40, cylinder end fill = black!60, opacity=0.4] (c) at (2.5,-3.3) {};
        
        \node[ellipse, draw, fill = gray!60, minimum width = 0.55cm, minimum height = 0.08cm] (prop11) at (0.1,1.15 - 0.45) {};
        \node[ellipse, draw, fill = gray!60, minimum width = 0.55cm, minimum height = 0.08cm] (prop12) at (1.0,1.15 - 0.45) {};
        \node[ellipse, draw, fill = gray!60, minimum width = 0.55cm, minimum height = 0.08cm] (prop13) at (1.0,1.7 - 0.45) {};
        \node[ellipse, draw, fill = gray!60, minimum width = 0.55cm, minimum height = 0.08cm] (prop14) at (0.1,1.7 - 0.45) {};
        \draw[line width=0mm] (prop11) -- (prop13);
        \draw[line width=0mm] (prop12) -- (prop14);
        
        \node[ellipse, draw, fill = gray!60, minimum width = 0.55cm, minimum height = 0.08cm] (prop21) at (1.05,-0.8 - 0.45) {};
        \node[ellipse, draw, fill = gray!60, minimum width = 0.55cm, minimum height = 0.08cm] (prop22) at (1.95,-0.8 - 0.45) {};
        \node[ellipse, draw, fill = gray!60, minimum width = 0.55cm, minimum height = 0.08cm] (prop23) at (1.8,-0.25 - 0.45) {};
        \node[ellipse, draw, fill = gray!60, minimum width = 0.55cm, minimum height = 0.08cm] (prop24) at (1.0,-0.25 - 0.45) {};
        \draw[line width=0mm] (prop21) -- (prop23);
        \draw[line width=0mm] (prop22) -- (prop24);
        
        \node[ellipse, draw, fill = gray!60, minimum width = 0.55cm, minimum height = 0.08cm] (prop31) at (1.05 + 3,-0.8 - 0.45) {};
        \node[ellipse, draw, fill = gray!60, minimum width = 0.55cm, minimum height = 0.08cm] (prop32) at (1.95 + 3,-0.8 - 0.45) {};
        \node[ellipse, draw, fill = gray!60, minimum width = 0.55cm, minimum height = 0.08cm] (prop33) at (1.8 + 3,-0.25 - 0.45) {};
        \node[ellipse, draw, fill = gray!60, minimum width = 0.55cm, minimum height = 0.08cm] (prop34) at (1.0 + 3,-0.25 - 0.45) {};
        \draw[line width=0mm] (prop31) -- (prop33);
        \draw[line width=0mm] (prop32) -- (prop34);
        
        \node[ellipse, draw, fill = gray!60, minimum width = 0.55cm, minimum height = 0.08cm] (prop11) at (0.1 + 3,1.15 - 0.45) {};
        \node[ellipse, draw, fill = gray!60, minimum width = 0.55cm, minimum height = 0.08cm] (prop12) at (1.0 + 3,1.15 - 0.45) {};
        \node[ellipse, draw, fill = gray!60, minimum width = 0.55cm, minimum height = 0.08cm] (prop13) at (1.0 + 3,1.7 - 0.45) {};
        \node[ellipse, draw, fill = gray!60, minimum width = 0.55cm, minimum height = 0.08cm] (prop14) at (0.1 + 3,1.7 - 0.45) {};
        \draw[line width=0mm] (prop11) -- (prop13);
        \draw[line width=0mm] (prop12) -- (prop14);
        
        

		\node at (2.5, -5.0) {\large $r_0, \ \pmb{R}_0$};
		
        \node at (3.5, 1.8) {\large $r_i, \ \pmb{R}_i$};
        
		\draw[-latex] (2.65, -3.3) -- (3.32, -2.7);
		\node at (2.65, -3.3)[circle,fill,inner sep=1.5pt]{};        
		\node at (2.6, -3.0) {\large $\rho_i$};
        
        \draw[<-] (-0.7, -5.2) -- (-1.2, -5.0);
        \draw[->] (-1.2, -5.0) -- (-1.2, -5.6);
        \draw[->] (-1.2, -5.0) -- (-0.8, -4.6);
        
        \node at (-0.6, -4.6) {\tiny x};
        \node at (-0.6, -5.2) {\tiny y};
        \node at (-1.3, -5.7) {\tiny z};
        
        \draw[-latex] (3.55, 0.95) -- (4.0, -0.1);
        \node at (4.2, 0.1) {$f_i$};
        
        \draw[{|}{latex}-{latex}{|}] (0.45, 0.95) -- (0.45, -2.1);
        \node at (0.3, -0.7) {$l_i$};
        
    \end{tikzpicture}
    
    \caption{A schematic diagram illustrating the payload transport using multiple UAVs. The support-frame together with the payload (shown lightly as a black cylinder) is modelled as a cuboid.}
    \label{fig:schematic_diagram_payloadUAV}
    
\end{figure}

Using the \textit{Lagrangian formulation} and the \textit{Lagrange-d'Alembert Principle}, one can arrive at the payload-UAV system dynamics:
\begin{gather}
    \begin{split}
        m_T\left(\dot{v}_0 + \omega_0^\times v_0\right) + \sum_{i=1}^N m_i \left( -\rho_i^\times\dot{\omega}_0 + \left(\omega_0^\times \right)^2\rho_i \right) \\ = \pmb{R}_0^T\left( m_Tg\mathbb{k} + \sum_{i=1}^N F_i \right), \ \ \ \ \ \ \label{eq:v_0}
    \end{split} \\
    \begin{split}
        \sum_{i=1}^Nm_i\rho_i^\times\left( \dot{v}_0 + \omega_0^\times v_0 \right) + \pmb{\Bar{J}}_0\dot{\omega}_0 + \omega_0^\times\pmb{\Bar{J}}_0\omega_0 \\ = \sum_{i=1}^N\rho_i^\times\pmb{R}_0^T\left( F_i + m_ig\mathbb{k} \right), \ \ \ \ \ \label{eq:w_0}
    \end{split} \\
    \begin{split}
        \pmb{J}_i\dot{\omega}_i + \omega_i^\times \pmb{J}\omega_i = \tau_i, \label{eq:w_i}
    \end{split}
\end{gather}
where the quantity $m_T = m_0 + \sum_{i=1}^Nm_i$ is the total combined mass, and $\pmb{\bar{J}}_0 = \left( \pmb{J}_0 - \sum_{i=1}^N m_i\left( \rho_i^\times \right)^2 \right)$ is the \textit{apparent moment of inertia} of the payload. For a detailed derivation, please refer to \cite{rao2022input}. In order to obtain a state-space representation of the payload-UAV dynamic model, define the state vector as:
\begin{align}
    x = \left[\underbrace{r_0^T \ v_0^T \ \Theta_0^T \ \omega_0^T}_\text{payload $\in \mathbb{R}^{12}$} \underbrace{\Theta_i^T \ \omega_i^T}_\text{$i^{th}$ UAV $\in \mathbb{R}^{6}$} \right]^T, \ i \in \{1,..., N \},
    \label{eq:state_def}
\end{align}
where $\Theta_0, \Theta_i\in\mathbb{R}^3$ are the $ZYX$ Euler angle characterization of the attitudes $\pmb{R}_0, \pmb{R}_i$ respectively. The nonlinear state-space equations can be obtained in the form $\dot{x} = f\left(x, u \right)$ by rearranging \eqref{eq:kin_beg} - \eqref{eq:w_i}, where the control input to the system is $u = \left[F_{t_i} \ \tau_{x_i} \ \tau_{y_i} \ \tau_{z_i} \right]^T \in \mathbb{R}^{4N}, \ i=1, \hdots, N$ and are given below (in addition to the kinematic equations of \eqref{eq:kin_beg} - \eqref{eq:kin_end}):
\begin{equation}
		\begin{bmatrix}
			\dot{v}_0 \\
			\dot{\omega}_0
		\end{bmatrix} = \pmb{\zeta} \times \pmb{\xi}, \label{eq:v0w0}
\end{equation}
where $\pmb{\zeta}\in\mathbb{R}^{6\times6}$, $\pmb{\xi}\in\mathbb{R}^{6\times1}$ and are given as follows:
\begin{equation}
    \pmb{\zeta} = \left[ \begin{array}{c|c}
         m_T\pmb{I}_3 & -\sum_{i=1}^N m_i\rho_i^\times \\
         \hline \sum_{i=1}^N m_i\rho_i^\times & \pmb{\bar{J}}_0
    \end{array}\right]^{-1} = \left[ 
    \begin{array}{c|c}
         \pmb{P}_{11} & \pmb{P}_{12} \\
         \hline
         \pmb{P}_{21} & \pmb{P}_{22}
    \end{array} \right] \label{eq:matrix_P}
\end{equation}
\begin{equation}\pmb{\xi} = \resizebox{.93\hsize}{!}{$
    \left[ \begin{array}{c}
         -m_{T} \omega_{0}^{\times} v_{0}-\sum_{i=1}^{N}m_{i}\left(\omega_{0}^{\times}\right)^{2} \rho_{i}+m_{T} g \pmb{R}_{0}^{T} \mathbb{k}+\sum_{i=1}^{N} \pmb{R}_{0}^{T} F_{i} \\ 
         \hline 
         -\omega_{0}^{\times} \Bar{\pmb{J}}_{0} \omega_{0}-\sum_{i=1}^{N} m_{i}\rho_{i}^{\times} \omega_{0}^{\times} v_{0}+\sum_{i=1}^{N} \rho_{i}^{\times} \pmb{R}_{0}^{T}\left(F_{i}+m_{i} g \mathbb{k}\right)
    \end{array}\right]$}
\end{equation}
In order to simplify the complexity of the system dynamics, the \textit{input-output feedback linearization} can be applied to the translational dynamics of the payload. Consider the output of the system to be the position of the payload, i.e., $y = r_0$. The \textit{relative degree} $\eta$ of the system is 2, since the control input appears in $\dot{v}_0$ as seen in \eqref{eq:v0w0}. Thus, one has to differentiate $r_0$ twice to obtain the control inputs explicitly:
\begin{align}
    \dot{r}_0 &= \pmb{R}_0v_0, \\
    \ddot{r}_0 &= \dot{\pmb{R}}_0v_0 + \pmb{R}_0\dot{v}_0, \label{eq:r_0_ddot}
\end{align}
where $\dot{v}_0$ is expanded from \eqref{eq:v0w0} as:
\begin{align}
    \begin{split}
    \dot{v}_0  &= \pmb{P}_{11}\left( -m_{T} \omega_{0}^{\times} v_{0}-\sum_{i=1}^{N}m_{i}\left(\omega_{0}^{\times}\right)^{2} \rho_{i}+m_{T} g \pmb{R}_{0}^{T} \mathbb{k} \right) \\ &+ \underbrace{\pmb{P}_{12}\left(  -\omega_{0}^{\times} \Bar{\pmb{J}}_{0} \omega_{0}-\sum_{i=1}^{N} m_{i}\rho_{i}^{\times} \omega_{0}^{\times} v_{0}+\sum_{i=1}^{N} \rho_{i}^{\times} \pmb{R}_{0}^{T}m_{i} g \mathbb{k} \right)}_\text{a function of state $\overline{f(x)}$} \\ &+ \underbrace{\pmb{P}_{11}\sum_{i=1}^{N} \pmb{R}_{0}^{T} F_{i} + \pmb{P}_{12}\sum_{i=1}^{N} \rho_{i}^{\times} \pmb{R}_{0}^{T}F_{i}}_\text{a function of state and control input}\label{eq:expand_v_0}
    \end{split} 
\end{align}
which can be simplified to:
\begin{align}
    \dot{v}_0 &= \overline{f(x)} + \left(\pmb{P}_{11}\pmb{S}_1 + \pmb{P}_{12}\pmb{S}_2\right)\overline{\mathbb{u}}, \label{eq:v_0_dot}
\end{align}
where $\pmb{S}_1 = \left[\pmb{I}_3 \ \hdots \ \pmb{I}_3 \right] \in \pmb{R}^{3\times 3N}$, $\pmb{S}_2 = \left[\rho_1^\times \ \hdots \ \rho_N^\times\right] \in \pmb{R}^{3\times 3N}$, and $\overline{\mathbb{u}} = \left[\mathbb{u}_1^T \ \cdots \ \mathbb{u}_N^T\right]\in\mathbb{R}^{3N}$ with $\mathbb{u}_i = \pmb{R}^T_0F_i$. Substituting \eqref{eq:v_0_dot} in \eqref{eq:r_0_ddot} yields:
\begin{align}
    \ddot{r}_0 &= \underbrace{\pmb{R}_0\omega_0^\times v_0 + \pmb{R}_0\overline{f(x)}}_\text{$d(x)$} + \underbrace{\pmb{R}_0\left(\pmb{P}_{11}\pmb{S}_1 + \pmb{P}_{12}\pmb{S}_2 \right)}_\text{$\Delta(x)$}\bar{\mathbb{u}}. \label{eq:r_b_delta}
\end{align}
Let $\overline{\mathbb{u}} = \Delta^\dagger(x)\mathbb{v} - d(x)$. Then, \eqref{eq:r_b_delta} simplifies to
\begin{align}
    \boxed{\ddot{y} = \ddot{r}_0 = \left[\begin{array}{l} \ddot{r}_{0_x} \\ \ddot{r}_{0_y} \\ \ddot{r}_{0_z} \end{array}\right] = \left[\begin{array}{l}\mathbb{v}_1 \\ \mathbb{v}_2 \\ \mathbb{v}_3 \end{array}\right] = \mathbb{v}.} \label{eq:fl_r0}
\end{align}
Thus, the feedback linearized dynamics of \eqref{eq:fl_r0} is a fairly simple linear system which is more conducive to the design of STL-based controllers. A more rigorous procedure of feedback linearization process for the multi-UAV payload system can be found in \cite{rao2022input}.
\subsection{Signal Temporal Logic (STL)}
Signal Temporal Logic (STL) \cite{maler2004monitoring} provides a formal framework to capture high-level specifications that can handle spatial, temporal, and logical constraints. It consists of a set of predicates $\mu$ that are evaluated based on their corresponding predicate function $h:\mathbb{R}^n \rightarrow \mathbb{R}$ as $\mu:=\left\{ \begin{array}{l} \text{True, \ \ \ \ \ if } h(x) \geq 0 \\ \text{False, \ \ \ \ if } h(x) < 0 \end{array} \right.$.
The syntax for an STL formula $\phi$ is given by:
\begin{align}
    \phi::=\text{True} \mid \mu \mid \neg\phi \mid \phi_1\wedge\phi_2 \mid \phi_1U_{[a, b]}\phi_2\mid F_{[a, b]}\phi \mid G_{[a, b]}\phi,
\end{align}
where $a, b \in \mathbb{R}^+_0$ with $a\leq b$, $\phi_1$ and $\phi_2$ are STL formulas, $F$ denotes the \textit{eventually} operator, $G$ denotes the \textit{always} operator and $U$ denotes the \textit{until} operator, each of which is defined below. The relation $(x, t)\models \phi$ indicates that the signal $x:\mathbb{R}^+_0 \rightarrow \mathbb{R}^n$ satisfies the STL formula $\phi$ at time $t$. 

\begin{definition}STL Semantics\cite{maler2004monitoring}:
The STL semantics for a signal $x$ are recursively defined as follows:
\begin{subequations}
\begin{alignat*}{2}
    &(x, t) \models \mu && \Leftrightarrow \ h(x) \geq 0 \\
    &(x, t) \models \neg \phi && \Leftrightarrow \neg\left( \left(x, t\right) \models \phi \right) \\
    &(x, t) \models \phi_1 \wedge \phi_2 && \Leftrightarrow (x, t) \models \phi_1 \wedge (x, t) \models \phi_2 \\
    &(x, t) \models \phi_1 U_{[a, b]}\phi_2 &&\Leftrightarrow \exists t_1 \in \left[t+a, t+b\right] s.t. (x, t_1)\models\phi_2 \\
    & &&\wedge \forall t_2\in\left[t, t_1\right], (x, t)\models\phi_1 \\
    &(x, t) \models F_{[a, b]}\phi && \Leftrightarrow \exists t_1 \in \left[t+a, t+b\right] s.t. (x, t_1) \models \phi \\
    &(x, t) \models G_{[a, b]}\phi &&\Leftrightarrow \forall t_1 \exists \left[t+a, t+b\right], (x, t_1)\models\phi.
\end{alignat*}
\end{subequations}
\end{definition}

\subsection{Higher-order Time-varying Control Barrier Functions}
Consider a nonlinear control-affine system:
\begin{align}
    \dot{x} = f(x) + g(x)u,
    \label{eq:nonlinear_affine}
\end{align}
where $x\in\mathbb{R}^n$ is the state of the system and $u\in\mathbb{R}^m$ is the control input to the system. Let the \textit{relative degree} of the system be $\eta > 1$. A time-varying set $\mathcal{C}(t)$ is \textit{forward invariant} if for a given control law $u$, there exists a unique solution $x(t)$ to the system  \eqref{eq:nonlinear_affine} with $x(t_0) = x_0 \in \mathcal{C}(t_0)$ such that $\forall t \in \left[t_0, t_f\right], x(t) \in \mathcal{C}(t)$.
For nonlinear affine systems  \eqref{eq:nonlinear_affine} with relative degree $\eta > 1$, \textit{higher order control barrier functions} (HOCBFs)\cite{xu2018constrained}\cite{xiao2021high} can be used to guarantee the \textit{forward invariance} of the set $\mathcal{C}(t)$.
\begin{definition} HOCBF(\cite{xiao2021high})
A function $b:\mathbb{R}^n\times\mathbb{R}^+ \rightarrow \mathbb{R}$ is a candidate higher order control barrier function (HOCBF) of relative degree $\eta$ for the system defined in \eqref{eq:nonlinear_affine} if there exists differentiable class $\mathcal{K}$ functions $\alpha_i, i \in \left\{1, 2, \hdots, \eta \right\}$ such that
\end{definition}
\begin{align}
    \begin{split}
    \sup_{u\in\mathcal{U}}\hspace{-0.2em} &\left[ \mathcal{L}^\eta_f b(x, t) \hspace{-0.2em} +\hspace{-0.2em}  \mathcal{L}_g\mathcal{L}_f^{\eta-1} b(x, t)u\hspace{-0.2em}  +\hspace{-0.2em}  \frac{\partial^\eta b(x, t)}{\partial t^\eta} \hspace{-0.2em} + \hspace{-0.2em} \alpha_\eta(\gamma_{\eta-1}\hspace{-0.2em} \left(x, t)\right )\right. \\ &\left. \left(\sum_{i=1}^{\eta-1} \mathcal{L}^i_f(\alpha_{\eta-i} \circ \gamma_{\eta-i-1}) + \frac{\partial^i\left( \alpha_{\eta-i} \circ \gamma_{\eta-i-1} \right)}{\partial t^i} \right) \right] \geq 0
    \end{split} \label{eq:hocbf}
\end{align}
where the functions $\gamma_i$ is defined as:
\begin{align}
    \gamma_i(x, t) = \dot{\gamma}_{i-1}(x, t) + \alpha_i\left(\gamma_{i-1}(x, t) \right), \ \ i = \{1, \hdots, \eta\}
\end{align}
with $\gamma_0$ chosen as $\gamma_0(x, t) = b(x, t)$.

\section{Temporal Waypoint Navigation of Multi-UAV Payload System}
\label{sec:TempNav}
\subsection{HOCBFs for STL Tasks}
This subsection discusses how higher-order time-varying control barrier functions can be used for STL tasks. Consider the following STL fragment:
\begin{align}
    \psi ::=& \ \text{True} \ | \ \mu \ | \ \neg\mu \ | \ \psi_1 \wedge \psi_2, \label{eq:psi}\\
    \phi ::=& \ G_{\left[a, b\right]}\psi \ | \ F_{\left[a, b\right]}\psi \ | \ \psi_1 U_{\left[a, b\right]}\psi_2 \ | \ \phi_1 \wedge \phi_2, \label{eq:phi}
\end{align}
where $\psi_1, \psi_2$ are STL formulas for class $\psi$ functions defined in \eqref{eq:psi} and $\phi_1, \phi_2$ are STL formulas for class $\phi$ functions defined in \eqref{eq:phi}.
The barrier functions are required to be less than or equal to their corresponding predicate function between the desired time interval. For tasks of the form $F_{[a,b]}\mu$ with predicate function $h(x)$, choose $b(x, t)$ such that $b(x, t') \leq h(x)$ for some $t' \in [a, b]$. For tasks of the form $G_{[a, b]}\mu$ with the predicate function $h(x)$, choose $b(x, t)$ such that $b(x, t') \leq h(x)$ for all $t' \in [a, b]$. When there are a conjunction of STL formulas as in \eqref{eq:psi} or \eqref{eq:phi}, the following smooth-approximation for minimum operator is used for a set of $M$ barrier functions with $j=\left\{1, \hdots, M\right\}$:
\begin{align}
    \min_{j} b_j(x, t) \approx -\text{ln}\left(\sum_j \text{exp}\left(-b_j(x, t)\right)\right).
\end{align}
Thus, for tasks of the form $F_{[a, b]}\psi$ where $\psi := \mu_1 \wedge \mu_2$ with the predicates $h_1(x)$ and $h_2(x)$ respectively, choose $b(x, t):= -\text{ln}\left( \text{exp}(-b_1(x, t)) + \text{exp}(-b_2(x, t)) \right)$ such that $b_1(x, t') \leq h_1(x)$ and $b_2(x, t') \leq h_2(x)$ for some $t' \in [a, b]$. Similarly, for tasks of the form $G_{[a, b]}\psi$ where $\psi := \mu_1 \wedge \mu_2$ with the predicates $h_1(x)$ and $h_2(x)$ respectively, choose $b(x, t):= -\text{ln}\left( \text{exp}(-b_1(x, t)) + \text{exp}(-b_2(x, t)) \right)$ such that $b_1(x, t') \leq h_1(x)$ and $b_2(x, t') \leq h_2(x)$ for all $t' \in [a, b]$ and so on. More details on using barrier functions for STL properties can be found in \cite{lindemann2018control}. 

Note that the relative degree for the feedback-linearized payload-UAV system is $\eta=2$. Thus, the control barrier function $b(x, t)$ which is a HOCBF, must satisfy \eqref{eq:hocbf} that can be simplified for $\eta=2$ as follows:
\begin{align}
    \begin{split}
    \underbrace{-\left(\mathcal{L}_g\mathcal{L}_fb\right)}_{P}u \leq & \ \ \mathcal{L}_f^2b + \frac{\partial^2b}{\partial t^2} + 2b\mathcal{L}_fb + 2b\frac{\partial b}{\partial t} + \left(\mathcal{L}_f b \right)^2 + \\ &\underbrace{\left( \frac{\partial b}{\partial t} \right)^2\hspace{-0.2em}+ b^4\hspace{-0.2em} + 2\left(\mathcal{L}_fb\right)\frac{\partial b}{\partial t} + \hspace{-0.2em}2b^2\mathcal{L}_fb + \hspace{-0.2em}2b^2\frac{\partial b}{\partial t}}_{H} \label{eq:HOCBF_constraint}
    \end{split}
\end{align}
where the class $\mathcal{K}$ functions $\alpha_i$'s are chosen as $\alpha_i(s) = s^2$ for all $s\in\mathbb{R}_0^+$ and $\gamma_0 = b$. Thus, the STL controller can be formulated as a \textit{convex quadratic programming} (CQP) problem as follow:
\begin{subequations}
\begin{align}
& \ \ \  \ \ \ \ \ \min_{u} \ \ u^T\pmb{Q}u \\
&\text{s.t.} \ \ \ \ \ \ \ \ \ Pu \leq H
\end{align}
\label{eq:CQP_defn}
\end{subequations}
At each time step, the state vector $x$ is obtained and the vectors $P, H$ are computed using \eqref{eq:HOCBF_constraint}, which results in a linear inequality $Pu \leq H$. Thus, at every time step, the CQP of \eqref{eq:CQP_defn} is solved to obtain the optimal control input $u$ to the system.

\subsection{Temporal Waypoint Navigation Problem}
Consider a series of $M$ waypoints. During the course of waypoint navigation, it is in general desirable that the payload reaches a particular waypoint location $\Lambda_j \in \mathbb{R}^3$ where $j=\left\{1, \hdots, M\right\}$ within a certain threshold radius $R_0 \in \mathbb{R}$, while meeting some time specifications before moving on to the next waypoint. This requires that the controller complies with both the \textit{spatial} as well as the \textit{temporal} requirement of the task at hand. For this purpose, a set of time-varying HOCBFs can be constructed to meet the required specifications. For the payload-UAV system, the structure of the HOCBFs is chosen as follows:
\begin{align}
    b_j\left(x, t\right) = \Gamma_j(t) - \left( \left\Vert r_0 - \Lambda_j \right\Vert_2 \right),
\end{align}
where $\Gamma_j$ is a \textit{temporal} function. For example, without any loss of generality, let the payload-UAV system start at the origin. The specifications to be met are as follows: $\phi := \phi_1 \wedge \phi_2$, where $\phi_1 := F_{[0, 14]}\left( \left\Vert r_0 - \left[ 2 \ 2 \ 2 \right]^T \right\Vert_2 \leq 0.1 \right)$ and $G_{\left[14, 25\right]}\left( \left\Vert r_0 - \left[ 2 \ 2 \ 2 \right]^T \right\Vert_2 \leq 3 \right)$. This means that the payload must reach the waypoint location $\Lambda = [2 \ 2 \ 2]^T$ within a radius of $R_0 = 0.1m$ in the time interval $t\in[0, 14]$. Once it reaches the specified waypoint, it must remain at the waypoint location $\Lambda = [2 \ 2 \ 2]^T$ within a radius of $R_0 = 3m$ in the time interval $t\in[14, 25]$. For this specification, the predicate function $h_1$ corresponding to $\phi_1$ is given by $h_1(x) = \left(0.1 - \left\Vert r_0 - \left[2\ 2\ 2 \right]^T \right\Vert_2\right)$. Similarly, the predicate function $h_2$ corresponding to $\phi_2$ is given by $h_2(x) = \left(3 - \left\Vert r_0 - \left[2 \ 2 \ 2 \right]^T \right\Vert_2\right)$. The barrier function $b_1$ corresponding to $\phi_1$ must satisfy the requirement $b_1(x, t) \leq h_1(x)$ for some $t' \in [0, 14]$. Similarly, the barrier function $b_2$ corresponding to $\phi_2$ must satisfy the requirement $b_2(x, t) \leq h_2(x)$ for all $t' \in [14, 25]$. The candidate barrier functions can be chosen as follows: 
\begin{subequations}
\begin{align}
b_1(x, t) &= \underbrace{\left(50.1 - \left(\frac{50}{14}\right)t\right)}_{\Gamma_1(t)} - \Big\Vert r_0 - \underbrace{\left[ 2 \ 2 \ 2 \right]^T}_{\Lambda_1} \Big\Vert_2 \\
b_2(x, t) &= \underbrace{\left( 347.93\cdot e^{-0.418t} + 2 \right)}_{\Gamma_2(t)} - \Big\Vert r_0 - \underbrace{\left[ 2 \ 2 \ 2 \right]^T}_{\Lambda_2}\Big\Vert_2 \\
b(x, t) &= -\text{ln}\left(e^{-b_1(x, t)} + e^{-b_2(x, t)}\right) \label{eq:final_barrier}
\end{align}
\end{subequations}
Note that $b_1(x, t) \leq h_1(x)$ for some $t' \in [0,14]$ and $b_2(x, t) \leq h_2(x)$ for all $t' \in [14, 25]$. 

\begin{figure*}
\centering
\includegraphics[height=5cm, width=17cm]{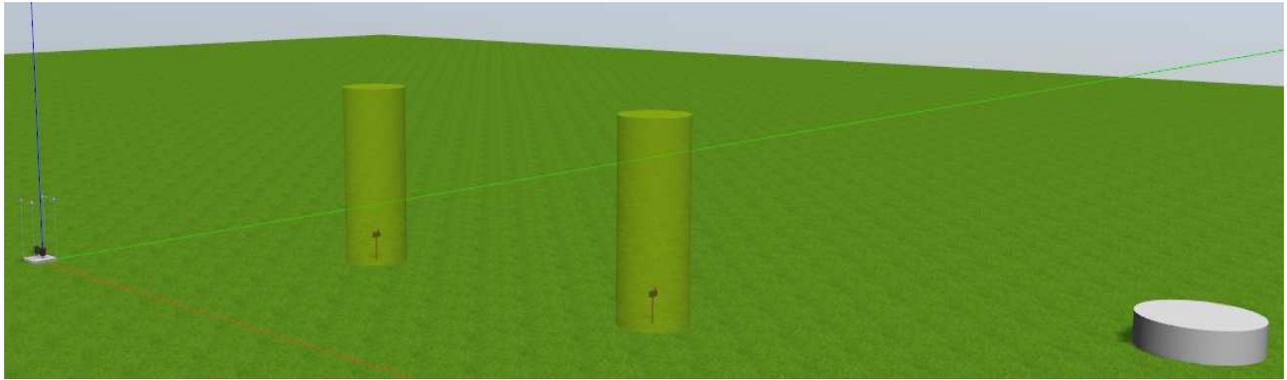}
\caption{Figure shows the simulation environment in the \texttt{Gazebo} simulator. The \textit{mailboxes} represent the location at which the package must be delivered. The multi-UAV payload system starts at the origin carrying two packages. Once both the packages are delivered, the payload must reach the hub (shown as a white cylindrical platform) for refilling purpose.}
\label{fig:env_gazebo}
\end{figure*}

This can then be extended to a series of $M$ waypoints that require both spatial as well as temporal specifications. The final barrier function $b(x, t)$ obtained from \eqref{eq:final_barrier} can then be plugged into \eqref{eq:HOCBF_constraint} to generate the HOCBF constraints. Finally, the CQP problem defined in \eqref{eq:CQP_defn} can be solved to obtain control inputs in online manner that ensures that the payload-UAV system meets the temporal waypoint navigation specifications.

\section{Simulation Results}
\label{sec:results}
In order to validate the proposed STL controller in \eqref{eq:CQP_defn}, a high-fidelity \texttt{Gazebo} simulation is conducted. The \texttt{rotorS MAV} simulator package \cite{Furrer2016} is used to spawn four UAVs that are connected to a rectangular payload via four rigid links. The software stack consists of custom \texttt{Gazebo} plugins that simulate GPS and IMU sensors. The entire payload-UAV model is spawned using \texttt{.xacro} files and the proposed control algorithm is implemented inside a \texttt{ROS C++} script. The control algorithm runs at a sampling rate of $50Hz$. The simulation is conducted on a 3.2GHz Intel i7 processor with 8GB RAM and NVIDIA 1050Ti graphics driver. 

\begin{figure}[!htb]
    \centering
    \includegraphics[scale=0.35]{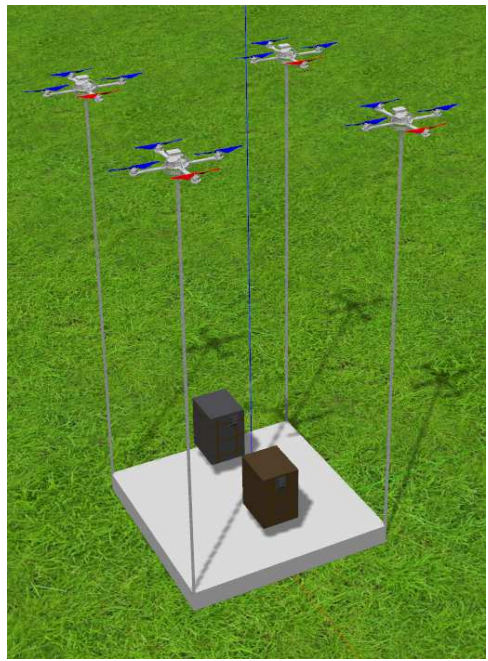}
    \caption{Figure shows 4 \textit{Hummingbird} UAVs connected to the vertical links through spherical joints. There are two packages on the payload, which the UAVs must transport to the required destination under certain time specifications.}
    \label{fig:multiUAV}
\end{figure}
The multi-UAV payload system is shown in Fig. \ref{fig:multiUAV}. It consists of four \textit{Hummingbird} UAVs that are connected to the four vertical rods via \textit{spherical joints} (modelled as two \textit{revolute joints} in \texttt{.xacro} file). The vertical rods are rigidly connected to the payload and thus, cannot move. The payload carries two packages that are to be delivered at specific locations with certain time specifications. 

\begin{table}
    \centering
    \caption{Parameters for the multi-UAV payload system}
    \label{tab:my_label}
    \begin{tabular}{ccccc} \toprule
         & $m$ & $J_{xx}$ & $J_{yy}$ & $J_{zz}$ \\ \midrule
         \text{Payload} & 1.0 & 0.556 & 0.556 & 0.556 \\
         \text{UAVs} & 0.68 & 0.029 & 0.029 & 0.055 \\ \midrule
         & & $\rho_i$ & & $l_i$ \\ \midrule
         \text{UAV1} & & $[0.25 \ \ \ \ \ 0.25 \ \ \ \ \ 0.125]^T$ & & 3.2 \\
         \text{UAV2} & & $[0.25 \ \ \ \ \shortminus0.25 \ \ \ \ 0.125]^T$ & & 3.2 \\
         \text{UAV3} & & $[\shortminus0.25 \ \ \ \shortminus0.25 \ \ \ 0.125]^T$ & & 3.2 \\
         \text{UAV4} & & $[\shortminus0.25 \ \ \ \ \ 0.25 \ \ \ \ \ 0.125]^T$ & & 3.2 \\\bottomrule
    \end{tabular}
\end{table}

\begin{figure*}[!htb]
    \centering
    \includegraphics[height=4.4cm]{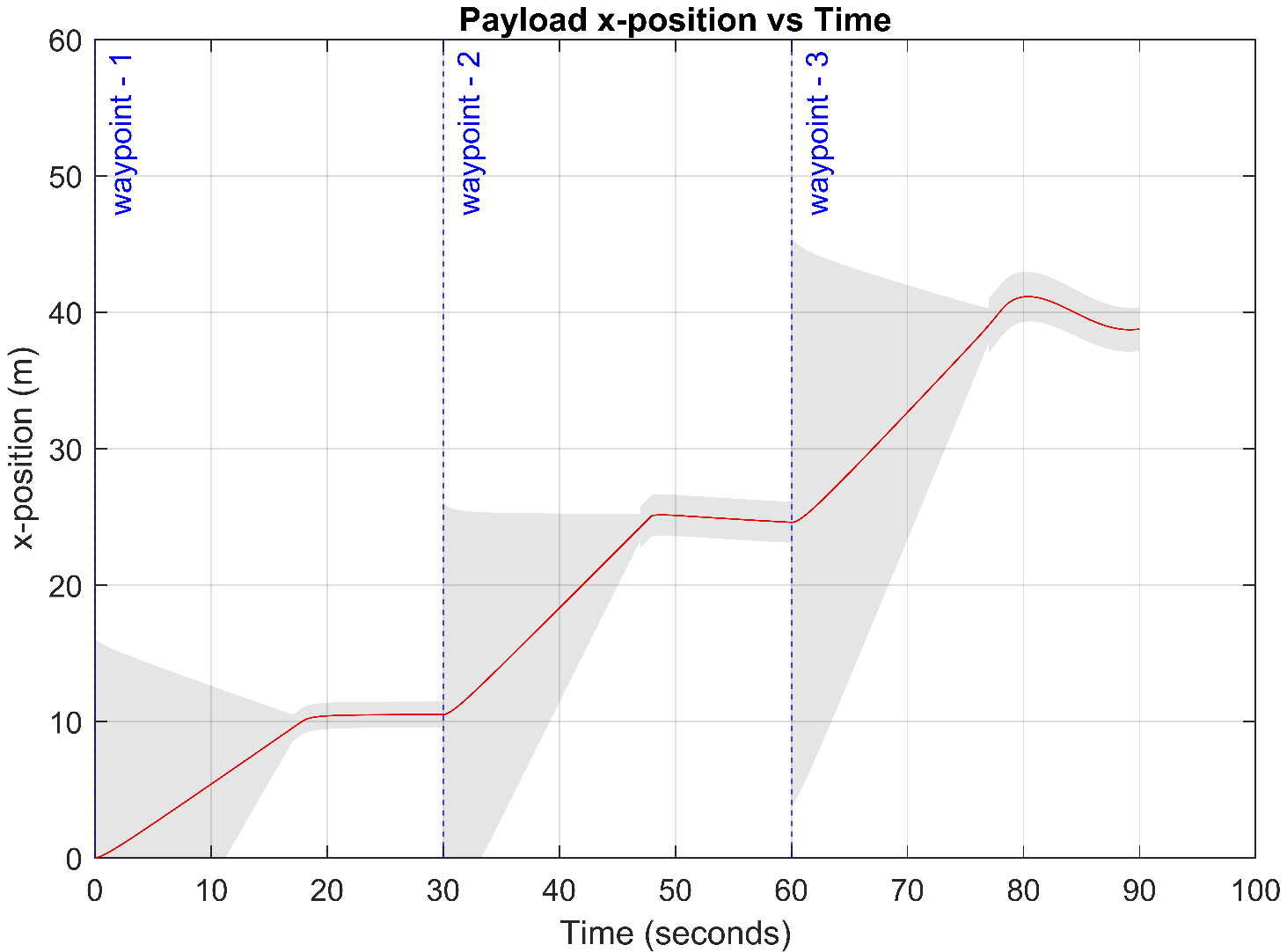}\includegraphics[height=4.4cm]{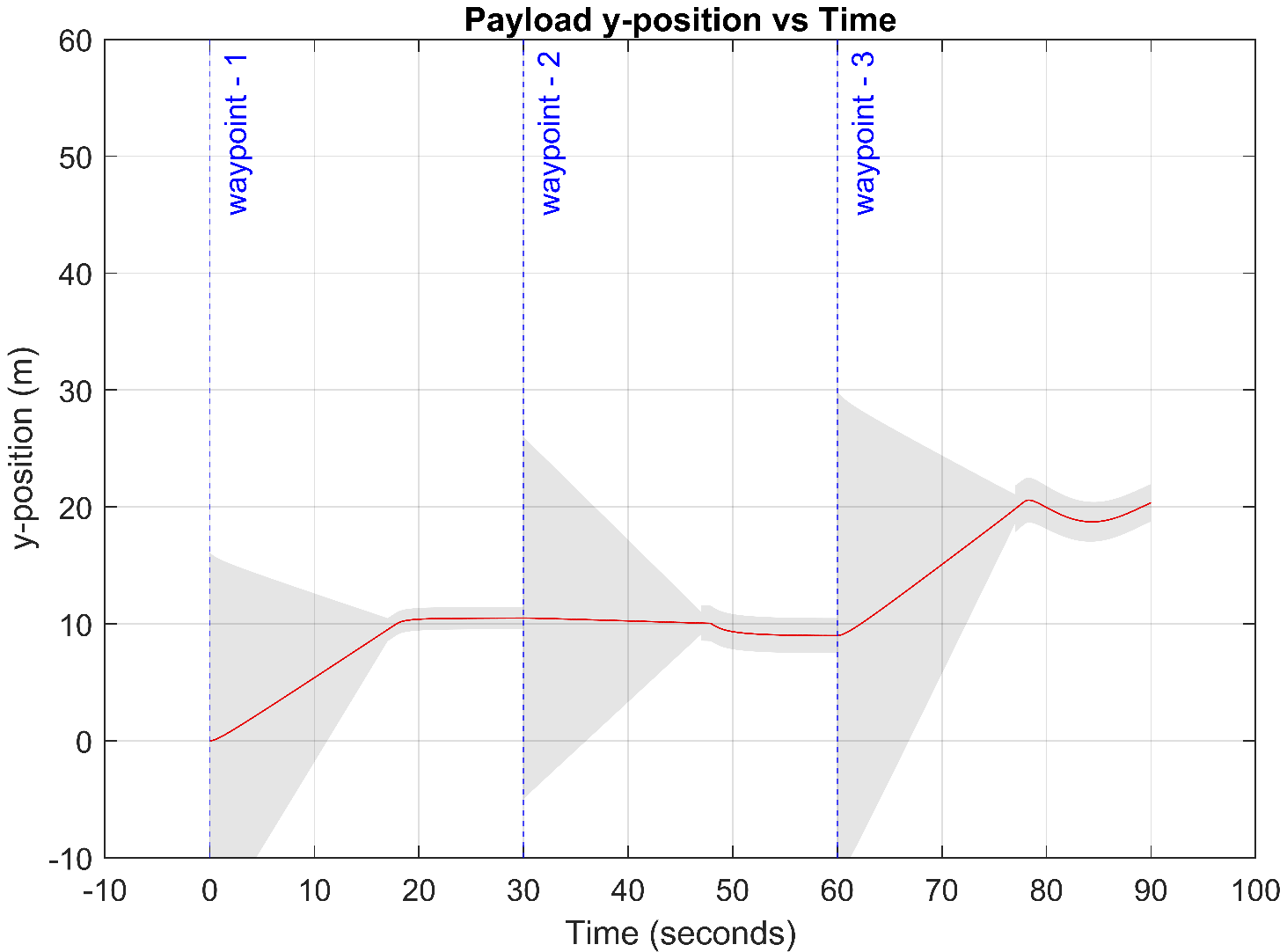}\includegraphics[height=4.4cm]{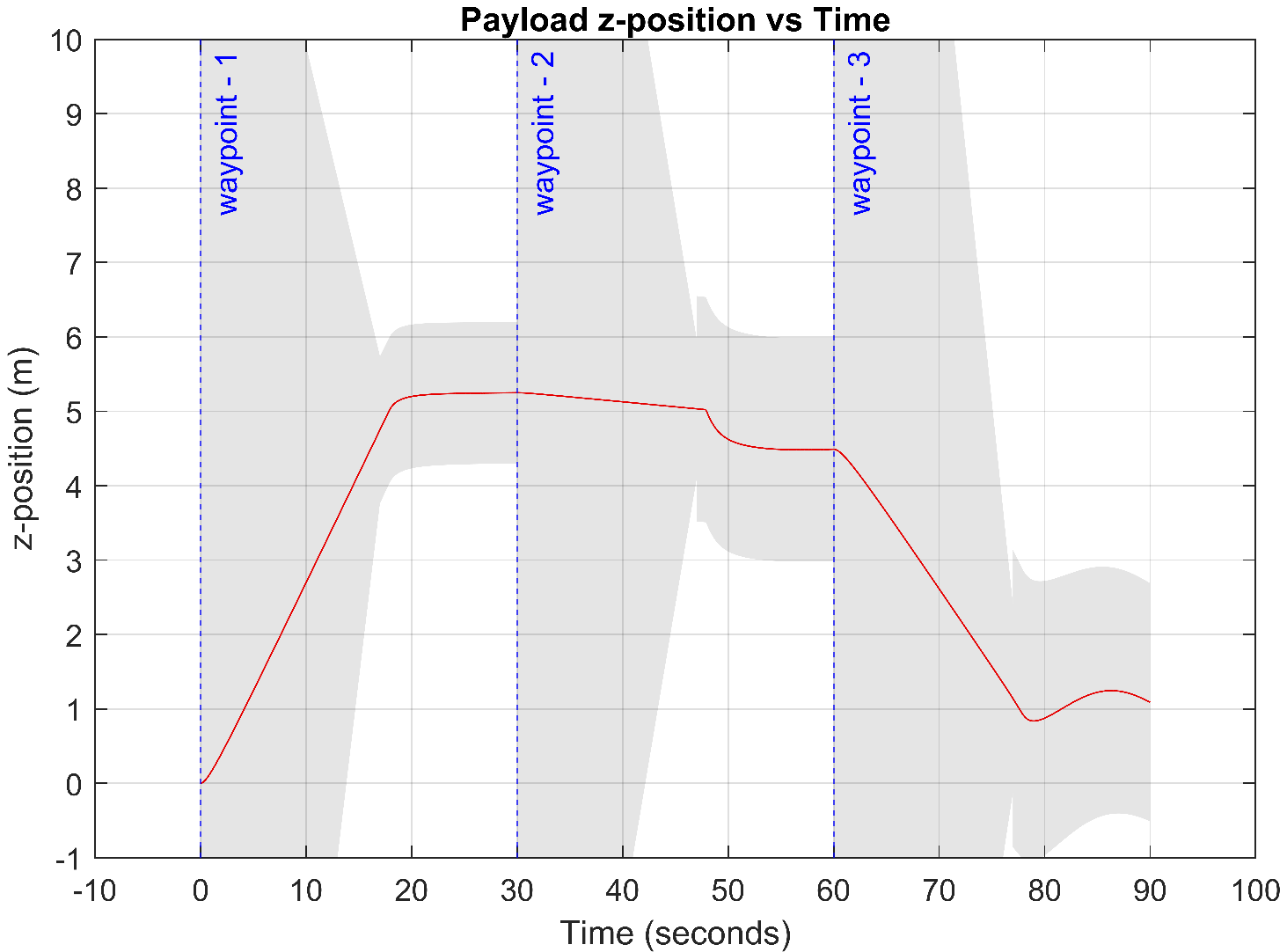}
    \caption{Figure shows the variation of $x$, $y$, and $z$ position of the payload with respect to time. The grey regions indicate the envelope of the corresponding temporal function.}
    \label{fig:traj}
\end{figure*}

Consider a typical package delivery scenario where the UAVs must collaboratively transport the payload carrying multiple packages from one point to another. For the sake of simulation purposes, consider that there are two packages on the payload that must be delivered at different locations with some temporal specifications. The payload must arrive at $\Lambda_1 = (10,10,5)$ to deliver the first package between the time interval $[0, 18]$ and stay there till $t=30s$ within a tolerance radius of $R_0 = 1.0$. The payload must then arrive at $\Lambda_2 = (25, 10, 5)$ to deliver the second package between the time interval $[30, 48]$ and stay there till $t=60s$ within the tolerance radius of $R_0 = 1.0$. Finally, the empty payload must land at the hub located at $\Lambda_3=(40, 20, 1)$ for refilling within the time interval $[60, 78]$ and stay there till $t=90s$ within the tolerance radius of $R_0 = 1.0$. These requirements can be captured in the following STL specification $\phi := \phi_1 \wedge \phi_2 \wedge \phi_3 \wedge \phi_4$ where,
\begin{align}
\begin{split}
    \phi_1 :=& F_{[0, 18]}\left(\left\Vert r_0 - \left[10 \ 10 \ 5 \right]^T \right\Vert_2 \leq 1.0\right) \wedge \ \\ & G_{[17, 30]}\left(\left\Vert r_0 - \left[10 \ 10 \ 5 \right]^T \right\Vert_2 \leq 1.0\right)
\end{split}
\end{align}
\begin{align}
\begin{split}
    \phi_2 :=& F_{[30, 48]}\left(\left\Vert r_0 - \left[25 \ 10 \ 5 \right]^T \right\Vert_2 \leq 1.0\right) \wedge \\ & G_{[47, 60]}\left(\left\Vert r_0 - \left[25 \ 10 \ 5\right]^T\right\Vert_2 \leq 1.0 \right)
\end{split}
\end{align}
\begin{align}
\begin{split}
    \phi_3 :=& F_{[60, 78]}\left(\left\Vert r_0 - \left[40 \ 20 \ 1\right]^T \right\Vert_2 \leq 1.0 \right) \wedge \\ & G_{[77, 90]}\left(\left\Vert r_0 - \left[40\ 20\ 1\right]^T \right\Vert_2 \leq 1.0\right)
\end{split}
\end{align}

\begin{align}
    \phi_4 := G_{[0, 90]}\left(\left\Vert r_0 \right\Vert_\infty \leq 50 \right) \ \ \ \label{eq:phi_4}
\end{align}
The specification $\phi_4$ in \eqref{eq:phi_4} ensures that the payload always remains inside a safe cuboid of side $50m$ centered at the origin.

\begin{figure}[!htb]
    \centering
    \includegraphics[height=5cm]{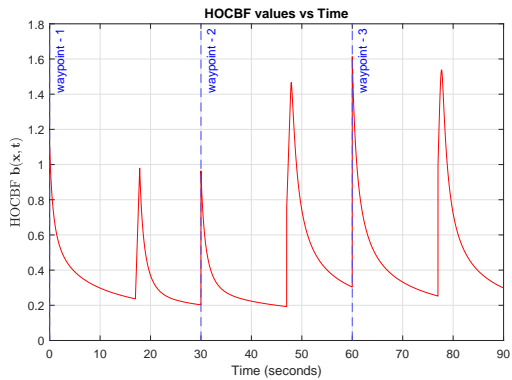}
    \caption{Figure shows the variation of the time-varying barrier function $b(x, t)$ with respect to time. }
    \label{fig:hocbf_t}
\end{figure}

The entire simulation environment is shown in Fig. \ref{fig:env_gazebo}. It consists of two \textit{mailboxes} positioned at $\Lambda_1$ and $\Lambda_2$ respectively. While delivering the packages, the payload's center of mass must be within the \textit{green translucent} cylinders that have a radius equal to $R_0$ which is the tolerance radius in the horizontal plane. Once delivered, the UAVs must land the payload on the white circular hub located at $\Lambda_3$. The simulation video is available \href{https://youtu.be/T1MYlKHaMkw}{here}\footnote{The video demonstration is available on this link if the pop-up doesn't work: \href{https://youtu.be/T1MYlKHaMkw}{https://youtu.be/T1MYlKHaMkw}}.

The payload trajectory in individual axis is shown in Fig. \ref{fig:traj} along with the time envelopes of the respective temporal functions $\Gamma_j(t)$. It can be seen that the controller satisfies all the temporal specifications while navigating through each of the waypoints. The values for time-varying barrier function $b(x, t)$ are provided in Fig. \ref{fig:hocbf_t}. 

\section{Conclusions}
\label{sec:future}
In this paper, the problem of \textit{temporal} waypoint navigation for multi-UAV payload systems is addressed. The $N$ UAVs are attached to rigid rods via spherical joints. These rods are connected to the payload rigidly and thus remain vertical always. This system's complex nonlinear state-space equations are simplified by applying input-output feedback linearization to the payload's translational dynamics. The Signal Temporal Logic Specifications are captured using higher-order time-varying control barrier functions to formulate an optimization-based control law. The controller's efficacy and real-time implementability are demonstrated by simulating a package delivery scenario in a high-fidelity \texttt{Gazebo} simulator. 
\bibliography{references}
\bibliographystyle{IEEEtran}

\end{document}